\documentclass[nohyperref]{article}

\usepackage{microtype}
\usepackage{graphicx}
\usepackage{subfigure}
\usepackage{multirow}
\usepackage{xcolor}

\usepackage{booktabs} 

\usepackage{hyperref}

\usepackage[accepted]{icml2023}

\usepackage{amsmath}
\usepackage{amssymb}
\usepackage{mathtools}
\usepackage{amsthm}

\usepackage[capitalize,noabbrev]{cleveref}

\theoremstyle{plain}

\theoremstyle{definition}

\theoremstyle{remark}

\usepackage[textsize=tiny]{todonotes}

\usepackage{siunitx}

\usetikzlibrary{decorations.pathreplacing,calligraphy}

\definecolor{fillbluelight}{RGB}{184, 187, 249}
\definecolor{fillblue}{RGB}{137, 142, 251}
\definecolor{lineblue}{RGB}{66, 74, 254}

\definecolor{fillgreenlight}{RGB}{184, 249, 195}
\definecolor{linegreen}{RGB}{66, 254, 129}

\definecolor{fillredlight}{RGB}{249, 184, 184}
\definecolor{fillred}{RGB}{251, 137, 137}
\definecolor{linered}{RGB}{254, 66, 66 }

\definecolor{lineyellow}{RGB}{254, 246, 66}

\definecolor{fillorangelight}{RGB}{249, 213, 184 }
\definecolor{fillorange}{RGB}{251, 187, 137}

\usepackage{stmaryrd}
\icmltitlerunning{E(n) Equivariant Message Passing Simplicial Networks}

\usetikzlibrary {arrows.meta}

\begin{document}

\twocolumn[
\icmltitle{E(n) Equivariant Message Passing Simplicial Networks}



\icmlsetsymbol{equal}{*}

\begin{icmlauthorlist}
\icmlauthor{Floor Eijkelboom}{uva}
\icmlauthor{Rob Hesselink}{uva}
\icmlauthor{Erik Bekkers}{uva}

\end{icmlauthorlist}

\icmlaffiliation{uva}{University of Amsterdam}

\icmlcorrespondingauthor{Floor Eijkelboom}{eijkelboomfloor@gmail.com}

\icmlkeywords{Machine Learning, ICML}

\vskip 0.3in
]



\printAffiliationsAndNotice{}  

\begin{abstract}
    This paper presents $\mathrm{E}(n)$ Equivariant Message Passing Simplicial Networks (EMPSNs), a novel approach to learning on geometric graphs and point clouds that is equivariant to rotations, translations, and reflections. 
    EMPSNs can learn high-dimensional simplex features in graphs (e.g. triangles), and use the increase of geometric information of higher-dimensional simplices in an $\mathrm{E}(n)$ equivariant fashion. 
    EMPSNs simultaneously generalize $\mathrm{E}(n)$ Equivariant Graph Neural Networks to a topologically more elaborate counterpart and provide an approach for including geometric information in Message Passing Simplicial Networks, thereby serving as a proof of concept for combining geometric and topological information in graph learning.
    The results indicate that EMPSNs can leverage the benefits of both approaches, leading to a general increase in performance when compared to either method individually, being on par with state-of-the-art approaches for learning on geometric graphs. Moreover, the results suggest that incorporating geometric information serves as an effective measure against over-smoothing in message passing networks, especially when operating on high-dimensional simplicial structures.
\end{abstract}

\section{Introduction}

The use of symmetry as an inductive bias in deep-learning models has been critical for the recent developments in areas such as drug repositioning \cite{zitnik2018modeling}, protein biology \cite{gligorijevic2021structure}, healthcare \cite{cosmo2020latent}, and traffic forecasting \cite{derrow2021eta}, among many others. One notable example of architectures exploiting symmetry constraints are graph neural networks (GNNs) \cite{scarselli2008graph}. Graphs lend themselves as useful descriptors for data that live on an irregular domain, such as molecules, meshes, or social networks.
One of the most common types of GNNs is Message Passing Neural Networks (MPNNs) \cite{gilmer2017neural}, where adjacent nodes send messages to each other and update their hidden representation accordingly. 

 MPNNs can be considered a differentiable and parameterized counterpart of the 1-dimensional Weisfeiler-Lehman test (1-WL) on graphs \cite{xu2018powerful}, where the features of the nodes describe the colors and adjacency relations are defined by the edges of the graph (see \citet{kipf2016semi}). As a consequence, MPNNs are at most as expressive as the 1-WL and will give equal predictions to non-isomorphic graphs that are not distinguished by 1-WL. Therefore, such models are limited in learning higher dimensional graph structures such as cliques, as also explored in \citet{chen2020can}.
One solution to this limitation is considering higher-dimensional simplices in the graph as learnable features \cite{bodnar2021weisfeiler}, thereby considering a topologically more elaborate space. Allowing for higher-dimensional features allows MPNNs to distinguish more graph isomorphisms than can be distinguished with the 1-WL test/standard MPNNs. The increased expressivity using higher-dimensional simplicial structures has also been explored in \citet{morris2019weisfeiler}. 

Many real-life problems have a natural symmetry to translations, rotations, and reflections (that is, to the Euclidean group $\mathrm{E}(n)$), such as object recognition or predicting molecular properties \cite{ramakrishnan2014quantum}. Many approaches leveraging these extra symmetries have been proposed, such as Tensor Field Networks \cite{thomas2018tensor}, $\mathrm{SE}(3)$ Transformers \cite{fuchs2020se}, $\mathrm{E}(n)$ Equivariant Graph Neural Networks \cite{satorras2021n}, among others. In contrast to using a more elaborate topology, these methods use the underlying geometry of the space in which the graph is positioned to improve expressivity. Even though these methods improve greatly from incorporating geometric information, they are still limited in their expressivity by not being able to explicitly learn higher-dimensional features explicitly present in the graph.

We present $\mathrm{E}(n)$ Equivariant Message Passing Simplicial Networks (EMPSNs), an $\mathrm{E}(n)$ equivariant formulation of Simplicial Message Passing Networks as introduced by \citet{bodnar2021weisfeiler}. This work serves as a proof of concept for combining geometric and topological graph approaches to leverage both benefits. We provide the following contributions:
\begin{itemize}
    \item We provide a generalization of $\mathrm{E}(n)$ Equivariant Graph Neural Networks (EGNNs) which can learn features on simplicial complexes. This approach incorporates more $\mathrm{E}(n)$ invariant information in the message passing procedure inspired by DimeNet-like architectures \cite{gasteiger2020directional}.
    \item We experimentally show that the use of higher-dimensional simplex learning improves performance compared to EGNNs and MPSNs, without requiring more parameters. This improvement is also found in datasets with few higher-dimensional simplices. We finally show that EMPSNs are competitive with state-of-the-art approaches on graphs, as illustrated in the N-body experiment and QM9. We also show that this improvement is obtained without a much greater forward time than other existing approaches. 
    \item We show that the performance of EMPSNs scales with the size and dimension of the simplicial complex, contrary to standard MPSNs. Moreover, the results indicate that incorporating geometric information is an effective approach to combating over-smoothing in graph networks in all dimensions. 
    
\end{itemize}

\section{Background}\label{sec:background}

In this section, we introduce the relevant definitions of equivariant and topological message passing. 

\paragraph{Equivariance}
In mathematics, the symmetries of an object are formalized using groups. Let $G$ be a group and let $\mathcal{X}$ and $\mathcal{Y}$ be sets on which a group action of $G$ is defined. 
A function $f: \mathcal{X} \to \mathcal{Y}$ is called \textbf{equivariant} to $G$ when $f$ commutes with the group action, i.e. when doing a transformation according to $g \in G$ and then evaluating the function gives the same result as first evaluating the function and then doing the transformation.
Such a function is called \textbf{invariant} to $G$ if computing the function on a transformed element yields the same outcome as computing the function in the untransformed element. 
Observe that invariance is therefore a specific type of equivariance. Formally, these properties can be described as
\begin{align*}
	\text{equivariance: } & f(g \cdot x) = g \cdot f(x),\\
	\text{invariance: } &f(g \cdot x) = f(x),
\end{align*}
for all $g \in G, x \in \mathcal{X}$, where $G$ acts both on the the input and output space. Enforcing equivariance in models has the advantage that no information will be lost when the model input is transformed, guaranteeing more stable predictions under predefined symmetries.

\paragraph{Message passing}
Message passing neural networks (MPNNs) are an influential class of graph networks proposed by \citet{gilmer2017neural}.
Let $\mathcal{G} = (\mathcal{V}, \mathcal{E})$ be a graph consisting of nodes $\mathcal{V}$ and edges $\mathcal{E}$. Suppose that each node $v_i \in \mathcal{V}$ and edge $e_{ij} \in \mathcal{E}$ has an associated node feature $\textbf{f}_i \in \mathbb{R}^{c_n}$ and edge feature $\textbf{a}_{ij} \in \mathbb{R}^{c_e}$ respectively, for some dimensionalities $c_n, c_e \in \mathbb{N}_{>0}$. In message passing, the hidden states of the nodes are iteratively updated by:
\begin{align*}
	\text{Find messages from $v_j$ to $v_i$} &:\hspace{3mm} \textbf{m}_{ij} = \phi_m(\textbf{f}_i, \textbf{f}_j, \textbf{a}_{ij}) \\
	\text{Aggregate messages to $v_i$} &:\hspace{3mm}\textbf{m}_{i} = \underset{j \in \mathcal{N}(i)}{\mathsf{Agg}}\textbf{m}_{ij} \\
	\text{Update hidden state $\textbf{f}_i$} &:\hspace{3mm} \textbf{f}'_i = \phi_f(\textbf{f}_i, \textbf{m}_i),
\end{align*}
where $\mathcal{N}(i)$ represents the set of neighbours of node $v_i$, the aggregation $\mathsf{Agg}$ is any permutation invariant function over the neighbours (e.g. summation), and $\phi_m$ and $\phi_f$ are commonly parameterised by multilayer perceptrons (MLPs).
To get a hidden state representing the entire graph, a permutation invariant aggregator is applied to all final hidden states of the nodes.

\paragraph{Equivariant message passing networks}
In some applications, the nodes in our graph are embedded in some Euclidean space forming a geometric graph. This spatial information can be incorporated in the message passing framework to account for physical information as seen in  \citet{thomas2018tensor}, \citet{fuchs2020se}, \citet{fuchs2021iterative}, \citet{klicpera2020directional}, \citet{gasteiger2021gemnet}, \citet{brandstetter2021geometric}.

A common model used on geometric graphs is the $\mathrm{E}(n)$ Equivariant Graph Neural Network (EGNN), which augments the message passing formulation to use the positional information while being equivariant to $\mathrm{E}(n)$ \cite{satorras2021n}. 
To exploit the geometric information in an $\mathrm{E}(n)$ equivariant fashion in message passing, the message function is conditioned on $\mathrm{E}(n)$ \textbf{invariant} information, e.g. the distance between two nodes. In the message passing framework, the first step is hence changed as follows:
$$\textbf{m}_{ij} = \phi_m (\textbf{f}_i, \textbf{f}_j, \mathsf{Inv}(\textbf{x}_i, \textbf{x}_j), \textbf{a}_{ij}),$$ for some function $\mathsf{Inv}$ that computes invariant attributes from the geometric quantities $\textbf{x}_i$ and $\textbf{x}_j$ in an $\mathrm{E}(n)$ invariant fashion, i.e.
$$\mathsf{Inv}(g \cdot \textbf{x}_i, g \cdot \textbf{x}_j) = \mathsf{Inv}(\textbf{x}_i, \textbf{x}_j),$$
for all $g \in \mathrm{E}(n).$ Moreover, in each layer the position of the nodes is updated in an equivariant fashion as follows:
$$\textbf{x}_i' = \textbf{x}_i + C \sum_{j \neq i} (\textbf{x}_i - \textbf{x}_j) \phi_x (\textbf{m}_{ij}),$$
for some MLP $\phi_x$ and constant $C$. This positional update is typically unused for $\mathrm{E}(n)$-invariant tasks such as predicting the internal energy of a molecule.

\begin{figure}[t!]
	\centering
	\begin{tikzpicture}[fill opacity=1, scale=1.5]
		\tikzstyle{point}=[circle,fill=black,inner sep=0pt,minimum size=3pt]
		\node (a)[point] at (0, .5) {};
		\node (b)[point] at (0.25, 0) {};
		\node (c)[point] at (0.75, 0.25) {};
		\node (d)[point] at (0.5, 0.75) {};
		\node (e)[point] at (0.75, 1) {};
		\node (f)[point] at (1, 0.75) {};
		\node (g)[point] at (1.25, 1.25) {};
		\node (h)[point] at (1.50, .50) {};
		\node (i)[point] at (1.25, .25) {};
		\node (j)[point] at (1.75, 0.25) {};
		\node (k)[point] at (1.5, 0.0) {};

		\draw[] (a.center) -- (b.center);
		\draw[] (a.center) -- (c.center);
		\draw[] (a.center) -- (d.center);
		\draw[] (b.center) -- (c.center);
		\draw[] (a.center) -- (b.center);
		\draw[] (b.center) -- (d.center);
		\draw[] (c.center) -- (d.center);
		\draw[] (d.center) -- (e.center);
		\draw[] (e.center) -- (f.center);
		\draw[] (d.center) -- (f.center);
		\draw[] (f.center) -- (g.center);
		\draw[] (e.center) -- (g.center);
		\draw[] (f.center) -- (h.center);
		\draw[] (f.center) -- (i.center);
		\draw[] (h.center) -- (i.center);
		\draw[] (c.center) -- (i.center);
		\draw[] (h.center) -- (j.center);
		\draw[] (k.center) -- (j.center);

		\draw[point] (a) circle (0.04);
		
		\draw[point] (b) circle (0.04);
		\draw[point] (c) circle (0.04);
		\draw[point] (d) circle (0.04);
		\draw[point] (e) circle (0.04);
		\draw[point] (f) circle (0.04);
		\draw[point] (g) circle (0.04);
		\draw[point] (h) circle (0.04);
		\draw[point] (i) circle (0.04);
		\draw[point] (j) circle (0.04);
		\draw[point] (k) circle (0.04);
		
		\begin{scope}[xshift=2.5cm]
			\node (a)[point] at (0, .5) {};
			\node (b)[point] at (0.25, 0) {};
			\node (c)[point] at (0.75, 0.25) {};
			\node (d)[point] at (0.5, 0.75) {};
			\node (e)[point] at (0.75, 1) {};
			\node (f)[point] at (1, 0.75) {};
			\node (g)[point] at (1.25, 1.25) {};
			\node (h)[point] at (1.50, .50) {};
			\node (i)[point] at (1.25, .25) {};
			\node (j)[point] at (1.75, 0.25) {};
			\node (k)[point] at (1.5, 0.0) {};

			\draw[fill=fillblue] (a.center) --  (b.center) -- (d.center) -- cycle;
			\draw[fill=fillbluelight] (c.center) --  (b.center) -- (d.center) -- cycle;
			\draw[fill=fillbluelight] (d.center) --  (e.center) -- (f.center) -- cycle;
			\draw[fill=fillbluelight] (g.center) --  (e.center) -- (f.center) -- cycle;
			\draw[fill=fillbluelight] (i.center) --  (h.center) -- (f.center) -- cycle;
			\draw (c.center) -- (i.center);
			\draw (h.center) -- (j.center);
			\draw (k.center) -- (j.center);
			\draw[dashed] (c.center) -- (a.center);

			\draw[point] (a) circle (0.04);
			
			\draw[point] (b) circle (0.04);
			\draw[point] (c) circle (0.04);
			\draw[point] (d) circle (0.04);
			\draw[point] (e) circle (0.04);
			\draw[point] (f) circle (0.04);
			\draw[point] (g) circle (0.04);
			\draw[point] (h) circle (0.04);
			\draw[point] (i) circle (0.04);
			\draw[point] (j) circle (0.04);
			\draw[point] (k) circle (0.04);
		\end{scope}

	\end{tikzpicture}
	\caption{Example of graph lifted to simplicial complex.} 
	\label{fig:graph_lift}
\end{figure}
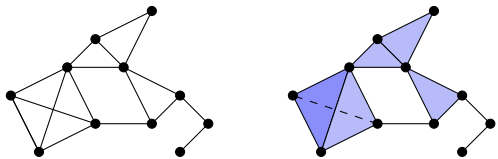

\paragraph{Simplicial complexes}\label{par:simcom} In geometry, a simplex is the generalization of triangles to other dimensions. Where a triangle ($2$-simplex) is formed by a set of $3$ fully connected points in space, an $n$-simplex is formed by a fully connected set of $n+1$ points. Examples of $n$-simplices are points (0-simplices), lines (1-simplices), and tetrahedra ($3$-simplices). 
To assign features to higher-dimensional simplices in our graph, a generalized notion of graphs called abstract simplicial complexes is considered.

An abstract simplicial complex (ASC) $\mathcal{K}$ is a collection of non-empty finite subsets of some set $\mathcal{V}$ such that for every set $\mathcal{T} \in \mathcal{K}$ and any non-empty subset $\mathcal{R} \subseteq \mathcal{T}$, it is the case that $\mathcal{R} \in \mathcal{K}$. In other words, an ASC is a set of simplices, such that any lower-dimensional simplex of the simplices in the ASC is also in the ASC. For example, if a triangle is part of the complex, then so are its sides and vertices. 
Though an ASC is a purely combinatorial object rather than a geometric one, it provides a natural way to associate a set of vertices of some graph $\mathcal{G}$ to a higher-order structure using a lifting transformation. The lifting transformation -- as illustrated in \autoref{fig:graph_lift} -- of a graph $\mathcal{G}$ is the ASC $\mathcal{K}$ with the property that if nodes $\{v_0, ..., v_k\}$ form a clique in $\mathcal{G}$, then the simplex $\{v_0, ..., v_k\} \in \mathcal{K}$ (see \citet{bodnar2021weisfeiler} for more information). By associating features to each simplex, one can use a more elaborate adjacency structure in the ASC to do message passing, as will be illustrated in the next section. Please note that a graph can be regarded as a 1-dimensional simplicial complex, where the nodes are the $0$-simplices and the $1$-simplices are the edges.

\paragraph{Message passing simplicial networks}
A simplex $\sigma$ is on the boundary of some simplex $\tau$, denoted $\sigma \prec \tau$, iff $\sigma \subset \tau$ and there exists no $\delta$ such that $\sigma \subset \delta \subset \tau.$ Observe that an $n$-dimensional simplex has $n+1$ boundaries if $n>0$, e.g. a triangle has its three sides as boundaries. Message Passing Simplicial Networks (MPSNs) as introduced in \citet{bodnar2021weisfeiler} provides a message passing framework in which more complex forms of adjacencies between objects in an ASC are considered. Specifically, the following adjacencies are distinguished:
\begin{enumerate}
	\item Boundary adjacencies $\mathcal{B}(\sigma) = \{\tau \mid \tau \prec \sigma \}$;
	\item Co-boundary adjacencies $\mathcal{C}(\sigma) = \{\tau \mid \sigma \prec \tau \}$;
	\item Lower-adjacencies $\mathcal{N}_{\downarrow}(\sigma) = \{\tau \mid \exists \delta, \delta \prec \tau \land \delta \prec \sigma \}$;
	\item Upper-adjacencies $\mathcal{N}_{\uparrow}(\sigma) = \{\tau \mid \exists \delta, \tau \prec \delta \land \sigma \prec \delta \}$.
\end{enumerate}

Please note that if our simplicial complex is a graph, the upper adjancencies of some node $v \in \mathcal{V}$ is simply the set of nodes $u$ that together with $v$ form an edge, i.e.

$$\mathcal{N}_{\uparrow}(v) = \{u \in \mathcal{V} \mid \{v, u\} \in \mathcal{E}\}.$$

Hence, standard GNNs communicate over the upper adjacencies of the nodes exclusively. 

Similarly to the standard message passing framework, the messages sent by the neighboring nodes are aggregated. For example, the boundary message to some simplex $\sigma$ is defined as follows:
\begin{equation*}
	\textbf{m}_{\mathcal{B}}(\sigma) = \underset{\tau \in \mathcal{B}(\sigma)}{\mathsf{Agg}}(\phi_{\mathcal{B}}(\textbf{f}_{\sigma}, \textbf{f}_{\tau})),
\end{equation*}
for some MLP $\phi_{\mathcal{B}}.$
Rather than incorporating just one message, there are four message types in the update:
\begin{equation*}
	\textbf{f}'_{\sigma} = \phi_f (\textbf{f}_{\sigma}, \textbf{m}_{\mathcal{B}}(\sigma), \textbf{m}_{\mathcal{C}}(\sigma), \textbf{m}_{\mathcal{N}_{\downarrow}}(\sigma), \textbf{m}_{\mathcal{N}_{\uparrow}}(\sigma)),
\end{equation*}
where $\textbf{m}_{\mathcal{B}}(\sigma), \textbf{m}_{\mathcal{C}}(\sigma), \textbf{m}_{\mathcal{N}_{\downarrow}}(\sigma),$ and $\textbf{m}_{\mathcal{N}_{\uparrow}}(\sigma)$ are the respective messages and  $\phi_f$ is the update MLP. In general, one can use different update and message MLPs for the different dimensional simplices. \citet{bodnar2021weisfeiler} also shows that this message passing framework is equally expressive when ignoring the co-boundaries and lower adjacencies. For a $k$ dimensional simplicial complex $\mathcal{K}$, the hidden state representing the entire complex, denoted by $\mathbf{h}_{\mathcal{K}}$, is found by concatenating simplex-invariant aggregation over the final hidden states:

$$\textbf{h}_{\mathcal{K}} := \bigoplus_{i=0}^k \underset{\substack{\sigma \in \mathcal{K}, \\ |\sigma| = i + 1}}{\mathsf{Agg}}\textbf{h}_\sigma,$$
where $\oplus$ denotes concatenation.

\section{E(n) Equivariant Message Passing Simplicial Networks}

$\mathrm{E}(n)$ Equivariant Message Passing Simplicial Networks (EMPSNs) generalize regular message passing neural networks to a $\mathrm{E}(n)$ equivariant counterpart on simplicial complexes. This is done in two steps:
\begin{enumerate}
	\item The input graph is lifted to a simplicial complex. This is done by either doing a graph lift or constructing a Vietoris–Rips complex.
	\item To each adjacency a set of $\mathrm{E}(n)$ invariant geometric attributes is assigned based on the communicating simplices. These so-called invariants are based on the positioning of the different points of the simplices in space.
\end{enumerate}
In this section, we will go over both steps and illustrate how the message passing formulation is altered. Note that the above two steps simply give us a set of neighborhoods for communication and geometric attributes between them. As such, our formulation is not restricted to any specific way of \textit{incorporating} geometric information, and thus any geometric graph approach could be used. In this work, we base our model on EGNN due to its simplicity and scalability, something on which we reflect a little more in the future research section.

\subsection{Defining the simplicial complex and adjacencies}

To leverage the higher-dimensional simplicial structure of the data, the input graph needs to be lifted to an ASC first. One of the aspects that make EGNNs highly effective is that EGNNs can operate on a fully connected graph and learn the relevance of each message passing connection based on the distance between the simplices, essentially defining an attention-like mechanism. Though in theory one could use a fully connected graph and assign a $n$-simplex to each clique of $n+1$ nodes, this approach would be hugely unscalable as a fully connected graph would have ${|\mathcal{V}|}\choose{n+1}$ simplices of dimension $n$, e.g. a fully connected small graph of $30$ nodes will have $435$ edges, $4060$ triangles, and $27,405$ tetrahedra.

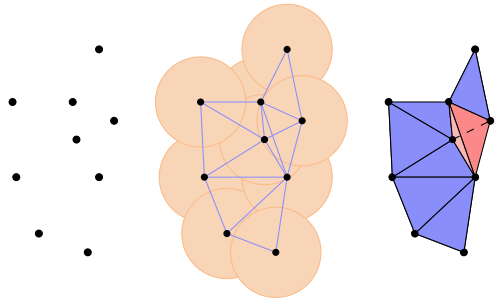
\begin{figure}[t]
	\centering
	\begin{tikzpicture}[fill opacity=0.5]
		\tikzstyle{point}=[circle,fill=black,inner sep=0pt,minimum size=3pt]
		\node (a)[point] at (.25,1.5) {};
		\node (b)[point] at (.55,.75) {};
		\node (c)[point] at (1.35,1.5) {};
		\node (d)[point] at (1.2,.5) {};
		\node (e)[point] at (1,2.5) {};
		\node (f)[point] at (1.05,2.0) {};
		\node (g)[point] at (1.35,3.2) {};
		\node (h)[point] at (1.55,2.25) {};
		\node (i)[point] at (0.2,2.5) {};

		\draw[point] (a) circle (0.04);
		\draw[point] (b) circle (0.04);
		\draw[point] (c) circle (0.04);
		\draw[point] (d) circle (0.04);
		\draw[point] (e) circle (0.04);
		\draw[point] (f) circle (0.04);
		\draw[point] (g) circle (0.04);
		\draw[point] (h) circle (0.04);
		\draw[point] (i) circle (0.04);
		

			

		
		\begin{scope}[xshift=2.5cm]
			\node (a)[point] at (.25,1.5) {};
			\node (b)[point] at (.55,.75) {};
			\node (c)[point] at (1.35,1.5) {};
			\node (d)[point] at (1.2,.5) {};
			\node (e)[point] at (1,2.5) {};
			\node (f)[point] at (1.05,2.0) {};
			\node (g)[point] at (1.35,3.2) {};
			\node (h)[point] at (1.55,2.25) {};
			\node (i)[point] at (0.2,2.5) {};
			
			\filldraw[color=fillorange, fill=fillorangelight] (.25,1.5) circle (0.6);
			\filldraw[color=fillorange, fill=fillorangelight] (.55,.75) circle (0.6);
			\filldraw[color=fillorange, fill=fillorangelight] (1.35,1.5) circle (0.6);
			\filldraw[color=fillorange, fill=fillorangelight] (1.2,.5)  circle (0.6);
			\filldraw[color=fillorange, fill=fillorangelight] (1,2.5) circle (0.6);
			\filldraw[color=fillorange, fill=fillorangelight] (1.05,2.0) circle (0.6);
			\filldraw[color=fillorange, fill=fillorangelight] (1.35,3.2) circle (0.6);
			\filldraw[color=fillorange, fill=fillorangelight] (1.55,2.25) circle (0.6);
			\filldraw[color=fillorange, fill=fillorangelight] (0.2,2.5) circle (0.6);

			\draw[fillblue] (a.center) --  (b.center);
			\draw[fillblue] (a.center) --  (c.center);
			\draw[fillblue] (a.center) --  (f.center);
			\draw[fillblue] (a.center) --  (i.center);
			
			\draw[fillblue] (b.center) --  (d.center);
			\draw[fillblue] (b.center) --  (c.center);

			\draw[fillblue] (c.center) --  (d.center);
			\draw[fillblue] (c.center) --  (e.center);
			\draw[fillblue] (c.center) --  (f.center);
			\draw[fillblue] (c.center) --  (h.center);
			
			\draw[fillblue] (e.center) --  (f.center);
			\draw[fillblue] (e.center) --  (g.center);
			\draw[fillblue] (e.center) --  (h.center);
			\draw[fillblue] (e.center) --  (i.center);
			
			\draw[fillblue] (f.center) --  (h.center);
			\draw[fillblue] (f.center) --  (i.center);
			\draw[fillblue] (g.center) --  (h.center);


			\draw[point] (a) circle (0.04);
			\draw[point] (b) circle (0.04);
			\draw[point] (c) circle (0.04);
			\draw[point] (d) circle (0.04);
			\draw[point] (e) circle (0.04);
			\draw[point] (f) circle (0.04);
			\draw[point] (g) circle (0.04);
			\draw[point] (h) circle (0.04);
			\draw[point] (i) circle (0.04);
		\end{scope}
		
		\begin{scope}[xshift=5cm]
			\node (a)[point] at (.25,1.5) {};
			\node (b)[point] at (.55,.75) {};
			\node (c)[point] at (1.35,1.5) {};
			\node (d)[point] at (1.2,.5) {};
			\node (e)[point] at (1,2.5) {};
			\node (f)[point] at (1.05,2.0) {};
			\node (g)[point] at (1.35,3.2) {};
			\node (h)[point] at (1.55,2.25) {};
			\node (i)[point] at (0.2,2.5) {};

			\draw (a.center) -- (b.center);
			\draw (b.center) -- (d.center);
			\draw (i.center) -- (a.center);
			\draw (d.center) -- (c.center);
			
			\draw[fill=fillblue] (a.center) --  (c.center) -- (f.center) -- cycle;
			\draw[fill=fillblue] (a.center) --  (i.center) -- (f.center) -- cycle;
			\draw[fill=fillblue] (b.center) --  (d.center) -- (c.center) -- cycle;
			\draw[fill=fillblue] (b.center) --  (a.center) -- (c.center) -- cycle;

			\draw[fill=fillblue] (h.center) --  (g.center) -- (e.center) -- cycle;
			\draw[fill=fillblue] (h.center) --  (g.center) -- (e.center) -- cycle;
			\draw[fill=fillblue] (i.center) --  (e.center) -- (f.center) -- cycle;
			\draw[fill=fillred] (c.center) --  (h.center) -- (e.center) -- cycle;
			\draw[fill=fillredlight] (c.center) --  (f.center) -- (e.center) -- cycle;
			
			\draw[dashed] (f.center) -- (h.center);

			\draw[point] (a) circle (0.04);
			\draw[point] (b) circle (0.04);
			\draw[point] (c) circle (0.04);
			\draw[point] (d) circle (0.04);
			\draw[point] (e) circle (0.04);
			\draw[point] (f) circle (0.04);
			\draw[point] (g) circle (0.04);
			\draw[point] (h) circle (0.04);
			\draw[point] (i) circle (0.04);
		\end{scope}
		
	\end{tikzpicture}
	\caption{Example of Vietoris Rips complex.} 
	\label{fig:vietorisrips}
\end{figure}

An alternative is constructing a simplicial complex based on the distances between the nodes similar to a radius graph. A common way for defining a simplicial complex as such is through constructing the Vietoris–Rips complex, which using some predefined distance $\delta$ contains a simplex for every set of points that lie at most a (Euclidean) distance $\delta$ away from each other. Formally, the Vietoris-Rips complex for some $\delta > 0$ - denoted $\mathsf{VietorisRips}(\delta)$ - is constructed by assigning a simplex so nodes $\{v_1, \cdots, v_n\} \subseteq \mathcal{V}$ if and only if $||\textbf{x}_i - \textbf{x}_j|| \leq \delta$ for all $0 \leq i,j \leq n$. This process is illustrated in \autoref{fig:vietorisrips}. In general, computing the Vietoris-Rips (VR) complex is exponential in the number of nodes, i.e. $\mathcal{O}(2^{|\mathcal{V}|})$. Hence, in general lifting to an ASC in each layer would add $L$
 such exponential operation in a model of 
 $L$ layers. In practise, this complexity will not be an issue, an argued in an analysis of the computational complexity and choice of construction of the ASC in \autoref{appendix:compcompl}.

An advantage of this approach over the standard graph lift illustrated in \autoref{sec:background} is that this approach enables higher-dimensional simplex learning on data that has few higher dimensional simplices, e.g. molecules. By increasing $\delta$ arbitrarily, we get a fully connected simplicial complex. This leads to a trade-off between the number of simplices in the complex and computation time during the message passing process. Similarly as done in EGNNs, we allow the model to learn the relative importance of each connection based on the geometric invariant defined over that adjacency. Note that as such the initial graph connectivity is lost, as it is when using EGNNs. Note that if this is undesired, it is always possible to do a regular graph lift to construct the ASC.

\subsection{Geometric information}
\label{geom_inf_def}

Given that the original graph is embedded in some Euclidean space, it is possible to condition the message passing function on $\mathrm{E}(n)$ invariant geometric information. The usage of a simplex to describe geometric information is natural, for considering the distances between set of $k$ points implicitly defines a $k-1$ dimensional simplex which geometry can be studied. In standard node-to-node communication, the defined simplex is an edge/line segment, and as such the only $\mathrm{E}(n)$ invariant attribute that can be considered is the length of the edge or any direct derivative of that length. When considering higher-dimensional simplices, however, there is more $\mathrm{E}(n)$ invariant information than a single distance one could leverage during message passing. The geometric information for the upper adjacent relations considered in this work is explored next.

\paragraph{Volumes} 
Let $\mathcal{K}$ be a simplicial complex embedded in $\mathbb{R}^n$ and let $\xi \in \mathcal{K}$ be a simplex in the complex. If the dimensionality of $\xi$ is greater than $0$, it is possible to assign a volume to $\xi$, denoted as $\mathsf{Vol}(\xi)$, defining a geometric invariant intrinsic to the feature. Hence, for each adjacency, the model is provided both the volume of the sending and receiving simplex. For some $n$-dimensional simplex $\xi = \{v_0, \cdots, v_n\}$ embedded in $\mathbb{R}^n$, its volume is given by
$$\mathsf{Vol}(\xi) = \frac{1}{n!}|~\text{det}\begin{pmatrix}v_1 - v_0 & \cdots & v_n - v_0\end{pmatrix}|.$$ Using Cayley–Menger determinants, also volumes of lower dimensional simplices in $\mathbb{R}^n$ can be computed, but these are not considered in this paper. 

\paragraph{Angles} Let $\sigma \in \mathcal{K}$ be an $n$-dimensional simplex and let $\tau, \eta \prec \sigma$ be distinct boundaries of $\sigma$. Since $\tau$ and $\eta$ are $(n-1)$-dimensional, they both define a hyperplane in $\mathbb{R}^n$ under the assumption that not all points of either boundary lie on a lower-dimensional plane, where the hyperplanes are denoted as $\mathrm{H}_\tau$, $\mathrm{H}_\eta$ respectively. This allows us to define the angle between $\tau$ and $\eta$, denoted as $\mathrm{Angle}(\tau, \eta)$ using the dihedral angle between their respective hyperplanes, i.e.
$$\mathrm{Angle}(\tau, \eta) := \arccos \frac{|\textbf{n}_\tau \cdot \textbf{n}_\eta|}{|\textbf{n}_\tau| |\textbf{n}_\eta|},$$
where $\textbf{n}_\tau, \textbf{n}_\eta$ denote the normal vectors of $\mathrm{H}_\tau$, $\mathrm{H}_\eta$. This process can be applied recursively to define new angles. Using $\tau$ as our point of reference, for two distinct boundaries $\delta_1, \delta_2 \prec \tau$, the boundaries define hyperplanes in $\mathrm{H}_\tau$, allowing us to define a new dihedral angle $\mathsf{Angle}(\delta_1, \delta_2)$ using their respective normals in $\mathrm{H}_\tau$. Note that this does not define an angle between $n$-dimensional objects in $\mathbb{R}^n$.

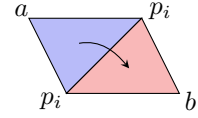
\begin{figure}[t]
	\centering
	\begin{tikzpicture}
		\tikzstyle{point}=[circle] 
		\node (a)[point] at (.5,1.5) {};
		\node (b)[point] at (1,.5) {};
		\node (c)[point] at (2,1.5) {};
		\node (d)[point] at (2.5,.5) {};

		\draw[fill=fillbluelight] (a.center) --  (b.center) -- (c.center) -- cycle;
		\draw[fill=fillredlight] (d.center) --  (b.center) -- (c.center) -- cycle;

		\draw[-stealth] (1.1667,1.1667) to [out=10,in=130] (1.8333,0.8333);

		\node[align=left] at (0.4,1.6) {$a$};
		\node[align=left] at (0.8,.4) {$p_i$};
		\node[align=left] at (2.25,1.6) {$p_i$};
		\node[align=left] at (2.65,.4) {$b$};

		
			
			
			
			
			
		
	\end{tikzpicture}
	\caption{Example of different invariants present in upper adjacent communication between $2$-simplices.} 
	\label{fig:invs}
\end{figure}

\paragraph{Distances} Comparable to EGNNs one can consider all invariants $||\textbf{x}_i - \textbf{x}_j||$ between the points of the simplices. To make sure that the final set of invariants is permutation invariant, we aggregate the relevant items in a permutation invariant fashion. Since two distinct upper adjacent simplices share all but one point, we can divide the relevant points into 1) the shared points $\{p_i\}$, 2) the unique point $a$ which is in the sending simplex but not in the receiving simplex, and 3) the unique point $b$ which is in the receiving simplex but not in the sending simplex. This division gives a way to generate a set of aggregates of geometric information in an invariant manner:

\begin{enumerate}
	\item Distances from the $p_i$ to $a$: $\mathsf{Agg}_i ~||\textbf{x}_{p_i} - \textbf{x}_a||$,
	\item Distances from the $p_i$ to $b$: $\mathsf{Agg}_i ~||\textbf{x}_{p_i} - \textbf{x}_b||$,
	\item Distances between the $p_i$: $\mathsf{Agg}_{i, j} ~||\textbf{x}_{p_i} - \textbf{x}_{p_j}||$,
	\item Distance from $a$ to $b$: $||\textbf{x}_a - \textbf{x}_b||$,
\end{enumerate}
as illustrated in \autoref{fig:invs}. These distances are then concatenated to form a $4$-dimensional invariant.

Whereas volume is an invariant that is both applicable for upper adjacent communication and boundary communication, both angles and distances are altered slightly for boundary communication.
For the invariant based on the boundary adjacency $\tau \prec \sigma$, we partition and aggregate the set of angles between all boundaries of $\sigma$ into 1) a set of angles with $\tau$ and 2) a set of angles without $\tau$ to define two aggregates. Moreover, since $\tau \subset \sigma$, there is no need for distances $(1)$ and $(4)$. Note that - even though not used in this work - there is no theoretical limitation to define invariant information based on equivariant features such as velocities in EGNN, e.g. for two velocities $\textbf{v}_i, \textbf{v}_j$ an invariant can be defined by taking their dot product.

\subsection{Equivariant simplicial message passing}

Message passing is done analogously to MPNNs, now conditioning each message function on the relevant $\mathrm{E}(n)$ invariant information based on the simplices that are communicated over. Let $\mathsf{Inv}(\sigma, \tau)$ denote the combined invariant as defined in \autoref{geom_inf_def}. Then, we simply find the aggregated message sent to some simplex $\sigma$ over a specific adjacency (e.g. boundaries) as follows:
$$\textbf{m}_{\mathcal{B}}(\sigma) = \underset{\tau \in \mathcal{B}(\sigma)}{\mathsf{Agg}}  \phi_{\mathcal{B}}(\textbf{f}_{\sigma}, \textbf{f}_{\tau}, \mathsf{Inv}(\sigma, \tau)).$$
We then update the features as done in the standard MPSN formulation. Since in some tasks we care about node predictions specifically, coboundary communication is always included - contrary to the only boundary and upper adjacent communication - such higher-dimensional geometric information can reach the nodes as well, allowing for geometrically stronger informed features on the nodes.

\begin{figure}[t]
	\centering
	\begin{tikzpicture}[fill opacity=1, scale=1.5]
		\tikzstyle{point}=[circle,fill=black,inner sep=0pt,minimum size=3pt]
		\node (a)[point] at (0, .65) {};
		\node (b)[point] at (0.15, .15) {};
		\node (c)[point] at (0.75, 0.4) {};
		\node (d)[point] at (0.5, 1) {};
		
		\draw[] (a.center) -- (b.center);
		\draw[] (a.center) -- (c.center);
		\draw[] (a.center) -- (d.center);
		\draw[] (b.center) -- (c.center);
		\draw[] (a.center) -- (b.center);
		\draw[] (c.center) -- (d.center);

            \draw[fill=fillbluelight] (a.center) --  (b.center) -- (c.center) -- cycle;

            \draw[fill=fillredlight] (a.center) --  (d.center) -- (c.center) -- cycle;
            
             \draw[-stealth, line width=0.25mm] (a.center) -- (0, .9);
              \draw[-stealth, line width=0.25mm] (b.center) -- (0.45, 0);
               \draw[-stealth, line width=0.25mm] (c.center) -- (0.95, 0.3);
                \draw[-stealth, line width=0.25mm] (d.center) -- (0.7, 0.9);

            \draw[dashed] (b.center) -- (d.center);
        
		\draw[point] (a) circle (0.04);
		\draw[point] (b) circle (0.04);
		\draw[point] (c) circle (0.04);
		\draw[point] (d) circle (0.04);

        		\begin{scope}[xshift=1.75cm]
\node (a)[point] at (0, .9) {};
		\node (b)[point] at (0.45, 0) {};
		\node (c)[point] at (0.95, 0.3) {};
		\node (d)[point] at (0.7, 0.9) {};
		
		\draw[] (a.center) -- (b.center);
		\draw[] (a.center) -- (c.center);
		\draw[] (a.center) -- (d.center);
		\draw[] (b.center) -- (c.center);
		\draw[] (a.center) -- (b.center);
		\draw[] (c.center) -- (d.center);

            \draw[fill=fillbluelight] (a.center) --  (b.center) -- (c.center) -- cycle;

            \draw[fill=fillredlight] (a.center) --  (d.center) -- (c.center) -- cycle;

            \draw[dashed] (b.center) -- (d.center);
        
		\draw[point] (a) circle (0.04);
		\draw[point] (b) circle (0.04);
		\draw[point] (c) circle (0.04);
		\draw[point] (d) circle (0.04);
  
        \end{scope}

	\end{tikzpicture}
	\caption{Change of invariants after position updates.} 
	\label{fig:movingstuff}
\end{figure}
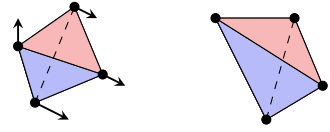

Furthermore, to ensure the geometry of the underlying simplices is consistent, only the positions of the nodes are updated. Please note that after each layer hence the invariant information between simplices might change, as illustrated in \autoref{fig:movingstuff}. When node positions are not updated - and hence the architecture is simply $\mathrm{E}(n)$ invariant - the model will be referred to as $\mathrm{E}(n)$ Invariant Message Passing Simplicial Networks (IMPSNs).

\section{Experiments}

For all experiments, the implementation and experimental details are provided in \autoref{appendix:implementation} and \autoref{appendix:expdetails} respectively.

\begin{table*}[t]

	\centering

	\label{tab:sample}
			\caption{Mean absolute error (MAE) on a subset of QM9 dataset. Relative improvement with respect to EGNNs (gain) is provided for comparison. \smallskip}
	\begin{tabular}{lccccccccccccc}
	\toprule
	Task & $\alpha$ & $\Delta \varepsilon$ & $\varepsilon_{\text{HOMO}}$ & $\varepsilon_{\text{LUMO}}$ & $\mu$ & $C_v$ & $G$ & $H$ & $R^2$ & $U$ & $U_0$ & \text{ZPVE} \\
	Units & bohr$^3$ & meV & meV & meV & D & cal/mol K & meV & meV & bohr$^3$ &meV & meV & meV\\
	\midrule
	 NMP & .092 & 69 & 43 & 38 & .030 & .040 & 19 & 17 & .180 & 20 & 20  &  1.50\\
	 SchNet * & .235 & 63 & 41 & 34 & .033 & .033 & 14 & 14 &.073 & 19 & 14 & 1.70\\
	 Cormorant & .085 & 61 & 34 & 38 & .038 & .026 & 20 & 21 & .961 & 21 & 22 & 2.02 \\
	 TFN & .223 & 58 & 40 & 38 & .064 & .101 & - & - & - & - & - &- \\
	 $\mathrm{SE}(3)$-Tr. & .142 & 53 & 35 & 33 & .051 & .054 & - & - & - & - & - &-\\
	 DimeNet++ * & \textbf{.043} & \textbf{32} & 24 & 19 & .029 & .023 & 7  &\textbf{6}& .331 & 6 & 6& 1.21  \\
	 SphereNet * & .046 & \textbf{32} & \textbf{23} & \textbf{18} & .026 & \textbf{.021} & 8 & \textbf{6}  &.292& 7 & 6 & 1.21\\
	 PaiNN * & .045 & 45 & 27 & 20 & \textbf{.012} & .024 & 7 & \textbf{6} &.066 & \textbf{5} & \textbf{5} &1.12\\
      SEGNN & .060 & 42 & 24 & 21 & .023 & .031 & 15 & 16 & .660 &12 & 15 &1.62\\
    MPSN & .266 & 153 & 89 & 77 & .101 &   .122 & 31 &  32 & .887 &33 & 33 &3.02 \\
	 EGNN & .071 & 48 & 29 & 25 & .029 & .031 & 12 & 12 & .106 &12 & 12 &1.55 \\
    \midrule
    \textbf{IMPSN} & .066 & 37 & 25 & 20 & .023 & .024 & \textbf{6} & 9  & .101 & 7 & 10& 1.37  \\
    \textbf{Gain} & 7\% & 23\% & 14\% & 20\% & 26\% & 23\% & 50\% & 25\%  & 4\% & 42\% & 17\% &12\%\\
    \bottomrule
	\end{tabular}

	    \label{tab:qm_results}
    \end{table*}

\paragraph{QM9}
\label{par:qm9}

The QM9 dataset \cite{ramakrishnan2014quantum} is a molecular dataset consisting of small molecules containing at most 29 atoms embedded in $3$-dimensional space. The task is to predict a series of chemical properties of the molecule. This dataset is especially interesting since only $\sim 43.7\%$ of molecules in this dataset contain a $2$-simplex, and hence performing a standard graph-lift would not leverage many of the benefits of MPSNs. Comparable to EGNN, the initial graph structure is dropped. The ASC constructed is combined with the standard fully connected graph to allow for a strictly more elaborate topological space. The results are provided in \autoref{tab:qm_results}. 

When comparing IMPSN to EGNN we observe that on all properties IMPSN outperforms EGNN, on average leading to an improvement of $22\%$ on those targets. Moreover, on many targets IMPSN performs almost on par with SOTA approaches on molecules, even beating SOTA in predicting free energy at 298.15K ($G$). This is an interesting result since the architecture is not curated for molecular tasks specifically, e.g. we do not leverage many of the molecule-specific intricacies such as Bessel function embeddings in our network as is done in  \citet{gasteiger2020directional}.

\paragraph{N-body system}

As introduced in \citet{kipf2018neural}, the N-body system experiment considers the trajectory in $3$-dimensional space of $5$ charged particles over time. The task is to predict the position of all bodies after 1,000 time steps, based on their initial positions and velocities. For a fair comparison, we use the experimental setup of this experiment as introduced in \citet{satorras2021n}. The results are summarized in \autoref{tab:res_nbody}. 

\begin{table}[h]
    \centering
    \caption{Mean Squared Error for the N-body system experiment. Average forward time for a batch size of 100 5-body systems is seconds is added for comparison. \\}
    \begin{tabular}{lcc}
        \toprule
        Method & MSE & Time (s) \\
        \midrule
        SE(3)-Tr. \cite{fuchs2020se} & .0244 & .1918 \\
        TFN \cite{thomas2018tensor} & .0155 & .0452 \\
        NMP \cite{gilmer2017neural} & .0107 & .0044 \\
        Radial Field \cite{kohler2019equivariant} & .0104 & .0049 \\
        SEGNN \cite{brandstetter2021geometric} & .0043 & .0672 \\
        MPSN \cite{bodnar2021weisfeiler} & .0808 & .0598 \\
        EGNN \cite{satorras2021n} & .0070 & .0158 \\
        \midrule
        \textbf{EMPSN} & .0063 & .0612 \\
         \bottomrule
    \end{tabular}
    \label{tab:res_nbody}
\end{table}

We observe a $10\%$ improvement over standard EGNNs when passing messages over the ASC, only being beaten by SEGNNs. This suggests that learning higher-order simplicial structures is beneficial to modeling N-body systems. Moreover, we observe that even though EMPSNs compute many more messages in each layer, the computational time does not exceed other approaches for N-body such as SEGNN.

\paragraph{EMPSN architecture and ablations} We compare standard MPSNs to EMPSNs to see how much MPSNs benefit from an increase in geometric information and how the improvement varies when we increase the dimensionality of the ASC. Also, by comparing MPSNs and EMPSNs over multiple dimensionalities, we simultaneously evaluate how much an increase in dimensionality improves the EGNN message-passing framework. These experiments hence implicitly define an ablation study for both the topological and geometric additions. 
Since the models with more elaborate simplicial communication require more parameters, we offset this by either 1) reducing the number of parameters in each MLP or 2) by giving the smaller models more layers to compensate. For both experiments, we used models with a fixed parameters budget ($\sim 200 \text{K}$). We report performance on the isotropic polarizability ($\alpha$) property of QM9 for ASC formed with radii of $\delta=\SI{3.0}{\angstrom}$ and $\delta=\SI{4.0}{\angstrom}$, where $\delta=3$ is the smallest $\delta$ value such that each molecule has at least one triangle. 

We call the highest dimensional adjacency relations used by the EMPSN the \textbf{type} of the EMPSN. For example, a (1-1) EMPSN is the EMPSN in which we do have all communication up to communication between 1-simplices and 1-simplices - and thus between 0-simplices and 0-simplices and 0-simplices and 1-simplices - but nothing higher. Similarly, a (1-2) EMPSN would have all the communication of the (1-1) EMPSN with added communication from 1-simplices to 2-simplices. The results for the different compensations are summarized in \autoref{tab:res_1} and \autoref{tab:res_2} respectively, where the improvement of EMPSNs relative to the MPSNs (or: gain) is also reported.

\begin{table}[h]
    \centering
    \caption{Mean absolute error (MAE) on QM9 $\alpha$ property using small models compensated with \textbf{number of hidden dimensions} ($\delta=\SI{3.0}{\angstrom}$ / $\SI{4.0}{\angstrom}$). Relative improvement when geometric information is given to the model (gain) is added for comparison. \\}
    \begin{tabular}{r|c|c|c}
        \toprule
        Type & MPSN & EMPSN & Gain \\
        \midrule
         0-0 & 0.188 / 0.310 & 0.118 / 0.107 & 1.6 / 2.9\\
         0-1 & 0.206 / 0.329 & 0.121 / 0.092 & 1.7 / 3.6 \\
        1-1 & 0.169 / 0.310 & 0.103 / 0.083 & 1.6 / 3.8 \\
         1-2 & 0.173 / 0.341 & 0.101 / 0.078 & 1.7 / 4.4 \\
         \bottomrule
    \end{tabular}
    \label{tab:res_1}
\end{table}

\begin{table}[h]
    \centering
    \caption{Mean absolute error (MAE) on QM9 $\alpha$ property using small models compensated with \textbf{number of layers} ($\delta=\SI{3.0}{\angstrom}$ / $\SI{4.0}{\angstrom}$). Relative improvement when geometric information is given to the model (gain) is added for comparison. \\}
    \begin{tabular}{r|c|c|c}
        \toprule
        Type & MPSN & EMPSN & Gain \\
        \midrule
         0-0 & 0.173 / 0.307 & 0.124 / 0.115 & 1.4 / 2.7\\
         0-1 & 0.195 / 0.323 & 0.133 / 0.100 & 1.5 / 3.3 \\
        1-1 & 0.165 / 0.296 & 0.107 / 0.085 & 1.5 / 3.5 \\
         1-2 & 0.173 / 0.341 & 0.101 / 0.078 & 1.7 / 4.4 \\
         \bottomrule
    \end{tabular}
    \label{tab:res_2}
\end{table}

In line with \citet{satorras2021n}, we see that providing the network with geometric information improves the performance of MPNNs. Moreover, we observe that increasing the connectivity of the graph improves performance on MPNNs when this geometric information is provided, but leads to a decrease in performance otherwise. This suggests that incorporating geometric information is an effective measure to combat overfitting. 

The results also suggest that increasing both the dimensionality and size of the ASC over which messages are passed indeed improves the performance of EMPSNs. However, when not using geometric information, there seems to be no added benefit to learning these higher-dimensional features, even leading to a decrease in performance when the ASC is too large. As we increase the dimensionality over which we do message passing, the gain increases dramatically, again supporting the claim that geometric information is an effective measure against over-smoothing. These effects persist both when we alter hidden dimensions and when we alter the number of message passing layers.


\section{Conclusion}

We presented EMPSNs, $\mathrm{E}(n)$ Equivariant Message Passing Simplicial Networks, a proof of concept on combining geometric and topological graph methods on geometric graphs and point clouds. These networks can learn using higher-dimensional simplices in graphs and take into account the increase in $\mathrm{E}(n)$ invariant geometric information during message passing. We provided a general approach for 1) lifting graphs to ASCs for message passing in a scalable fashion and 2) defining $\mathrm{E}(n)$ invariant information based on the relative positions of the communicating simplices. Our results indicate that the usage of higher dimensional emergent simplex learning is beneficial without requiring more parameters, hence leveraging the benefits of topological and geometric  methods. Moreover, the results indicated that our formulation is on par with SOTA approaches for learning on graphs. Last, we showed that using geometric information combats over-smoothing and that this effect is stronger in higher dimensions.

\paragraph{Limitations}

One limitation of this approach is the increased time complexity of the higher dimensional message passing approaches since we pass many more messages in each layer. An avenue worth exploring is allowing the model to pass messages over the different dimensional simplices, but not perform each message passing type in each layer, e.g. first pass messages from nodes to triangles in the first layers, and then pass messages over triangles only. This would cut down the time complexity of the model, while still leveraging its benefits.

\paragraph{Future work}

Since our work serves as a proof of concept for combining topological and geometric graph approaches, there are a lot of directions for future work that could be worthwhile to study.

An interesting direction for future research is considering more elaborate topological spaces than simplicial complexes and using the increased $\mathrm{E}(n)$ invariant information present to improve performance further. Especially in the case of molecular predictions, using topological spaces such as CW complexes could be hugely beneficial since those structures can model rings explicitly. Similarly, the usage of other geometric graph approaches can be combined with topological graph approaches. Moreover, two-hop connections and the corresponding invariants can be considered to generalize the framework, e.g. through concatenating the different invariants.

Moreover, since steerable methods based on Clebsch-Gordan products work really well by being able to capture aspects of the geometric graph such as relative orientation, it would be interesting to explore steerable alternatives to EGNN - e.g. SEGNN - as our base model to do message passing on. Even though we think the invariant approach on simplicial complexes is useful given that we can good performance without needing to go to equivariant features, we think that modeling topology explicitly and using equivariant features would leverage the best of both worlds if one can find a way to keep computational costs manageable.

Last, the EMPSN framework might lend itself to a general classification for many graph methods. It could be worthwhile to explore how different existing methods on graphs compare using our framework, possibly formulating our framework as general group convolutions on point clouds.

\bibliography{empsn}

\begin{thebibliography}{25}
\providecommand{\natexlab}[1]{#1}
\providecommand{\url}[1]{\texttt{#1}}
\expandafter\ifx\csname urlstyle\endcsname\relax
  \providecommand{\doi}[1]{doi: #1}\else
  \providecommand{\doi}{doi: \begingroup \urlstyle{rm}\Url}\fi

\bibitem[Bodnar et~al.(2021)Bodnar, Frasca, Wang, Otter, Montufar, Lio, and
  Bronstein]{bodnar2021weisfeiler}
Bodnar, C., Frasca, F., Wang, Y., Otter, N., Montufar, G.~F., Lio, P., and
  Bronstein, M.
\newblock Weisfeiler and lehman go topological: Message passing simplicial
  networks.
\newblock In \emph{International Conference on Machine Learning}, pp.\
  1026--1037. PMLR, 2021.

\bibitem[Brandstetter et~al.(2021)Brandstetter, Hesselink, van~der Pol,
  Bekkers, and Welling]{brandstetter2021geometric}
Brandstetter, J., Hesselink, R., van~der Pol, E., Bekkers, E., and Welling, M.
\newblock Geometric and physical quantities improve e (3) equivariant message
  passing.
\newblock \emph{arXiv preprint arXiv:2110.02905}, 2021.

\bibitem[Chen et~al.(2020)Chen, Chen, Villar, and Bruna]{chen2020can}
Chen, Z., Chen, L., Villar, S., and Bruna, J.
\newblock Can graph neural networks count substructures?
\newblock \emph{Advances in neural information processing systems},
  33:\penalty0 10383--10395, 2020.

\bibitem[Cosmo et~al.(2020)Cosmo, Kazi, Ahmadi, Navab, and
  Bronstein]{cosmo2020latent}
Cosmo, L., Kazi, A., Ahmadi, S.-A., Navab, N., and Bronstein, M.
\newblock Latent-graph learning for disease prediction.
\newblock In \emph{International Conference on Medical Image Computing and
  Computer-Assisted Intervention}, pp.\  643--653. Springer, 2020.

\bibitem[Derrow-Pinion et~al.(2021)Derrow-Pinion, She, Wong, Lange, Hester,
  Perez, Nunkesser, Lee, Guo, Wiltshire, et~al.]{derrow2021eta}
Derrow-Pinion, A., She, J., Wong, D., Lange, O., Hester, T., Perez, L.,
  Nunkesser, M., Lee, S., Guo, X., Wiltshire, B., et~al.
\newblock Eta prediction with graph neural networks in google maps.
\newblock In \emph{Proceedings of the 30th ACM International Conference on
  Information \& Knowledge Management}, pp.\  3767--3776, 2021.

\bibitem[Fuchs et~al.(2020)Fuchs, Worrall, Fischer, and Welling]{fuchs2020se}
Fuchs, F., Worrall, D., Fischer, V., and Welling, M.
\newblock Se (3)-transformers: 3d roto-translation equivariant attention
  networks.
\newblock \emph{Advances in Neural Information Processing Systems},
  33:\penalty0 1970--1981, 2020.

\bibitem[Fuchs et~al.(2021)Fuchs, Wagstaff, Dauparas, and
  Posner]{fuchs2021iterative}
Fuchs, F.~B., Wagstaff, E., Dauparas, J., and Posner, I.
\newblock Iterative se (3)-transformers.
\newblock In \emph{International Conference on Geometric Science of
  Information}, pp.\  585--595. Springer, 2021.

\bibitem[Gasteiger et~al.(2020)Gasteiger, Gro{\ss}, and
  G{\"u}nnemann]{gasteiger2020directional}
Gasteiger, J., Gro{\ss}, J., and G{\"u}nnemann, S.
\newblock Directional message passing for molecular graphs.
\newblock \emph{arXiv preprint arXiv:2003.03123}, 2020.

\bibitem[Gasteiger et~al.(2021)Gasteiger, Becker, and
  G{\"u}nnemann]{gasteiger2021gemnet}
Gasteiger, J., Becker, F., and G{\"u}nnemann, S.
\newblock Gemnet: Universal directional graph neural networks for molecules.
\newblock \emph{Advances in Neural Information Processing Systems},
  34:\penalty0 6790--6802, 2021.

\bibitem[Gilmer et~al.(2017)Gilmer, Schoenholz, Riley, Vinyals, and
  Dahl]{gilmer2017neural}
Gilmer, J., Schoenholz, S.~S., Riley, P.~F., Vinyals, O., and Dahl, G.~E.
\newblock Neural message passing for quantum chemistry.
\newblock In \emph{International conference on machine learning}, pp.\
  1263--1272. PMLR, 2017.

\bibitem[Gligorijevi{\'c} et~al.(2021)Gligorijevi{\'c}, Renfrew, Kosciolek,
  Leman, Berenberg, Vatanen, Chandler, Taylor, Fisk, Vlamakis,
  et~al.]{gligorijevic2021structure}
Gligorijevi{\'c}, V., Renfrew, P.~D., Kosciolek, T., Leman, J.~K., Berenberg,
  D., Vatanen, T., Chandler, C., Taylor, B.~C., Fisk, I.~M., Vlamakis, H.,
  et~al.
\newblock Structure-based protein function prediction using graph convolutional
  networks.
\newblock \emph{Nature communications}, 12\penalty0 (1):\penalty0 1--14, 2021.

\bibitem[Kingma \& Ba(2014)Kingma and Ba]{kingma2014adam}
Kingma, D.~P. and Ba, J.
\newblock Adam: A method for stochastic optimization.
\newblock \emph{arXiv preprint arXiv:1412.6980}, 2014.

\bibitem[Kipf et~al.(2018)Kipf, Fetaya, Wang, Welling, and
  Zemel]{kipf2018neural}
Kipf, T., Fetaya, E., Wang, K.-C., Welling, M., and Zemel, R.
\newblock Neural relational inference for interacting systems.
\newblock In \emph{International conference on machine learning}, pp.\
  2688--2697. PMLR, 2018.

\bibitem[Kipf \& Welling(2016)Kipf and Welling]{kipf2016semi}
Kipf, T.~N. and Welling, M.
\newblock Semi-supervised classification with graph convolutional networks.
\newblock \emph{arXiv preprint arXiv:1609.02907}, 2016.

\bibitem[Klicpera et~al.(2020)Klicpera, Gro{\ss}, and
  G{\"u}nnemann]{klicpera2020directional}
Klicpera, J., Gro{\ss}, J., and G{\"u}nnemann, S.
\newblock Directional message passing for molecular graphs.
\newblock \emph{arXiv preprint arXiv:2003.03123}, 2020.

\bibitem[K{\"o}hler et~al.(2019)K{\"o}hler, Klein, and
  No{\'e}]{kohler2019equivariant}
K{\"o}hler, J., Klein, L., and No{\'e}, F.
\newblock Equivariant flows: sampling configurations for multi-body systems
  with symmetric energies.
\newblock \emph{arXiv preprint arXiv:1910.00753}, 2019.

\bibitem[Loshchilov \& Hutter(2016)Loshchilov and Hutter]{loshchilov2016sgdr}
Loshchilov, I. and Hutter, F.
\newblock Sgdr: Stochastic gradient descent with warm restarts.
\newblock \emph{arXiv preprint arXiv:1608.03983}, 2016.

\bibitem[Morris et~al.(2019)Morris, Ritzert, Fey, Hamilton, Lenssen, Rattan,
  and Grohe]{morris2019weisfeiler}
Morris, C., Ritzert, M., Fey, M., Hamilton, W.~L., Lenssen, J.~E., Rattan, G.,
  and Grohe, M.
\newblock Weisfeiler and leman go neural: Higher-order graph neural networks.
\newblock In \emph{Proceedings of the AAAI conference on artificial
  intelligence}, volume~33, pp.\  4602--4609, 2019.

\bibitem[Ramakrishnan et~al.(2014)Ramakrishnan, Dral, Rupp, and von
  Lilienfeld]{ramakrishnan2014quantum}
Ramakrishnan, R., Dral, P.~O., Rupp, M., and von Lilienfeld, O.~A.
\newblock Quantum chemistry structures and properties of 134 kilo molecules.
\newblock \emph{Scientific Data}, 1, 2014.

\bibitem[Satorras et~al.(2021)Satorras, Hoogeboom, and Welling]{satorras2021n}
Satorras, V.~G., Hoogeboom, E., and Welling, M.
\newblock E (n) equivariant graph neural networks.
\newblock In \emph{International Conference on Machine Learning}, pp.\
  9323--9332. PMLR, 2021.

\bibitem[Scarselli et~al.(2008)Scarselli, Gori, Tsoi, Hagenbuchner, and
  Monfardini]{scarselli2008graph}
Scarselli, F., Gori, M., Tsoi, A.~C., Hagenbuchner, M., and Monfardini, G.
\newblock The graph neural network model.
\newblock \emph{IEEE transactions on neural networks}, 20\penalty0
  (1):\penalty0 61--80, 2008.

\bibitem[Tancik et~al.(2020)Tancik, Srinivasan, Mildenhall, Fridovich-Keil,
  Raghavan, Singhal, Ramamoorthi, Barron, and Ng]{tancik2020fourier}
Tancik, M., Srinivasan, P., Mildenhall, B., Fridovich-Keil, S., Raghavan, N.,
  Singhal, U., Ramamoorthi, R., Barron, J., and Ng, R.
\newblock Fourier features let networks learn high frequency functions in low
  dimensional domains.
\newblock \emph{Advances in Neural Information Processing Systems},
  33:\penalty0 7537--7547, 2020.

\bibitem[Thomas et~al.(2018)Thomas, Smidt, Kearnes, Yang, Li, Kohlhoff, and
  Riley]{thomas2018tensor}
Thomas, N., Smidt, T., Kearnes, S., Yang, L., Li, L., Kohlhoff, K., and Riley,
  P.
\newblock Tensor field networks: Rotation-and translation-equivariant neural
  networks for 3d point clouds.
\newblock \emph{arXiv preprint arXiv:1802.08219}, 2018.

\bibitem[Xu et~al.(2018)Xu, Hu, Leskovec, and Jegelka]{xu2018powerful}
Xu, K., Hu, W., Leskovec, J., and Jegelka, S.
\newblock How powerful are graph neural networks?
\newblock \emph{arXiv preprint arXiv:1810.00826}, 2018.

\bibitem[Zitnik et~al.(2018)Zitnik, Agrawal, and Leskovec]{zitnik2018modeling}
Zitnik, M., Agrawal, M., and Leskovec, J.
\newblock Modeling polypharmacy side effects with graph convolutional networks.
\newblock \emph{Bioinformatics}, 34\penalty0 (13):\penalty0 i457--i466, 2018.

\end{thebibliography}
\bibliographystyle{icml2023}

\newpage
\appendix
\onecolumn



\section{Constructing the abstract simplicial complexes}
\label{appendix:compcompl}

In this section, the design choices and computational complexity of constructing the ASCs are described. 

\subsection{\v{C}ech and Vietoris-Rips complexes}

When considering the topology formed by considering the union of $\delta$-balls around all points $v \in \mathcal{V}$, the case can be made that \v{C}ech complexes of size $\delta$ - denoted $\mathsf{\check{C}ech}(\delta)$ - more directly resemble this intuitive topology on the data. This is especially the case since $\mathsf{\check{C}ech}(\delta)$ is homotopy equivalent to this $\delta$-ball topological space - and the fact that part of the motivation of using simplicial complexes in DL comes from using persistent homologies. However, even though the runtime complexity of constructing a \v{C}ech complex is equal to that of constructing a Vietoris-Rips complex - both being $\mathcal{O}(2^{|\mathcal{V}|})$ - in practice, a Vietoris-Rips complex is much cheaper to compute. The main reason is that essentially constructing a Vietoris-Rips is implemented as 1) constructing a radius graph, and then 2) finding the cliques in that graph to do a graph lift. For instance, in practice, it is much cheaper to find all $5$ cliques in a graph than it is to loop over all subsets of size $5$ of the points as would be needed in constructing the \v{C}ech complex, as there you assign a simplex to set of vertices $\{v_1, \cdots, v_n\}$ iff $\bigcap_{i=1}^n U_{v_n} \neq \emptyset$, for some open set $U_{v_n}$ around $v_n$. Moreover, a formal case can be made that nothing is lost when choosing a Vietoris-Rips complex, since for any $\delta$ it holds that
$$\mathsf{\check{C}ech}(\delta) \subset \mathsf{VietorisRips}(\delta) \subset \mathsf{\check{C}ech}(2 \delta),$$
where $\mathsf{VietorisRips}(\delta)$ denotes the Vietoris-Rips complex of size $\delta$. That is, if we can find some $\delta$ such that the data is well described by the respective \v{C}ech complexes, then so will it be by our Vietoris-Rips complex.

\subsection{Computational efficiency}

In general, computing the Vietoris-Rips (VR) complex is exponential in the number of nodes, i.e. $\mathcal{O}(2^{|\mathcal{V}|})$. Hence, in general lifting to an ASC in each layer would add $L$
 such exponential operation in a model of 
 $L$ layers. However, often this exponential runtime is not an issue.

The first scenario where this forms no issue is when the ASC can be precomputed. In many situations, e.g. in QM9, it is common to not update the node positions in the model, in which case these ASCs can be computed before training. The second scenario where no problems are encountered is when a fully connected ASC can be used. This is seen in N-bdoy, as the number of nodes is small enough for one to use a fully connected ASC, i.e. for $5$ bodies one has 
 $10$ edges, $10$ triangles, and $5$ tetrahedra.

Only in the case where we do update positions and recompute the ASC in each layer, we do arrive at the exponential complexity. This is similar to how in EGNN one would have to recompute a radius graph when working with a larger geometric graph. However, in practice, for the Vietoris-Rips complexes, there is no need to do `intersection checks' over all subsets, and as such the complexity of this operation will be much more efficient than exponential.  Moreover, given that the radius we choose is fairly small in practice, one also saves a lot of computation when detecting cliques since our graph is sparse. To illustrate the point that indeed the runtime is limited, we computed the average runtime in ms for constructing the radius graph versus the Vietoris-Rips complex for different values of $\delta$
 on graphs in QM9. The implementations used are those of Pytorch Geometric and Gudhi respectively. The results in \autoref{tab:eff} clearly show that indeed we are a tiny bit slower for large values of 
$\delta$, but the difference is fairly subtle, especially in absolute terms. 

\begin{table}[h]
    \centering
    \begin{tabular}{c|c|c}
        $\delta ~(\SI{}{\angstrom})$ & $\mathsf{RadiusGraph}$ (ms) &  $\mathsf{VietorisRips}$ (ms)  \\
        \hline
        4 & .121 & .122 \\
        8 & .124 & .254 \\
        12 & .124 & .259 \\
        16 & .124 & .259 \\
        20 & .125 & .261 
    \end{tabular}
    \caption{Average runtime in ms for constructing the radius graph versus the Vietoris-Rips complex for different values of $\delta$ 
 on graphs in QM9. The implementations used are those of Pytorch Geometric and Gudhi respectively.}
    \label{tab:eff}
\end{table}

\section{Implementation}
\label{appendix:implementation}

In this section, we describe the implementation of EMPSNs. After having constructed the Vietoris-Rips complex from a geometric graph or point cloud, we have access to 1) a set of features for all different simplices plus initial features $\{\textbf{h}^n_i\}$ where $\textbf{h}^n_i$ denotes the $i^{\text{th}}$ feature of dimension $n$, 2) a set of adjacency relationships between these simplices, 3) positional information of the nodes $\{\textbf{x}_i\}$, and 4) optionally initial velocities of the nodes $\{\textbf{v}_i\}$. The main learnable functions are the following:

\begin{itemize}
    \item The features are embedded using linear embedding: $\text{Initial Feature} \to \{\text{LinearLayer}\} \to \text{Embedded Feature}$.
    \item For each adjacency, we learn a message function, e.g. for the adjacency $\mathcal{A}$ from $\tau$ to $\sigma$: $[\textbf{h}_\sigma, \textbf{h}_\tau, \mathsf{Inv}(\sigma, \tau)] \to \{\textbf{h}_\sigma \oplus \textbf{h}_\tau \oplus \mathsf{Inv}(\sigma, \tau) \to \text{LinearLayer} \to \text{Swish} \to \text{LinearLayer} \to \text{Swish} \} \to \textbf{m}^{\mathcal{A}}_{\sigma, \tau}$, where $\oplus$ denotes concatenation.
    \item For each message, an edge importance is computed: $\textbf{m}_\sigma^{\mathcal{A}} \to \{ \text{LinearLayer} \to \text{Sigmoid} \} \to e_{\sigma}^{\mathcal{A}}$
    \item For each simplex type, we learn an update for the simplex based on the different adjacency updates, i.e. $[\textbf{h}_\sigma, \{\textbf{m}_\sigma^{\mathcal{A}}\}, \{e_\sigma^{\mathcal{A}}  \}] \to \{\textbf{h}_\sigma \oplus (\bigoplus_{\mathcal{A}} e_{\sigma}^{\mathcal{A}} \cdot \textbf{m}_\sigma^{\mathcal{A}}) \to \text{LinearLayer} \to \text{Swish} \to \text{LinearLayer} \to \text{Addition}(\textbf{h}_\sigma) \} \to \textbf{h}_\sigma'$.
    \item A final readout: $\{\textbf{h}^n_i\} \to \{ \text{LinearLayer} \to \text{Swish} \to \text{LinearLayer} \to \bigoplus_n (\sum_i~ \textbf{h}^n_i) \to  \text{LinearLayer} \to \text{Swish} \to \text{LinearLayer}\} \to \text{Prediction}$.
\end{itemize}

These learnable functions are the same across all experiments. In all experiments, we included boundary, co-boundary, and upper adjacent communication. Moreover, if the initial graph has velocities, we update the position using two MLPs similar to done in \citet{satorras2021n}: 

\begin{itemize}
    \item A new velocity is computed based using two MLPs $\textbf{v}_i = \phi_v(\textbf{h}_i) \textbf{v}^{\text{init}}_i  + C \sum_{j \neq i} (\textbf{x}_i - \textbf{x}_j) \phi_x (\textbf{m}_{ij})$.
    \item The position is updated using the velocity, $\textbf{x}_i' =\textbf{x}_i + \textbf{v}_i$.
\end{itemize}

Both $\phi_v$ and $\phi_x$ are two-layer MLPs with a Swish activation function, i.e. $\text{Input} \to \{\text{LinearLayer} \to \text{Swish} \to \text{LinearLayer} \} \to \text{Output}$. Our experiments showed that not regressively updating $\textbf{v}_i$ in each layer but rather predicting $\textbf{v}_i$ based on the initial velocity $\textbf{v}_i^{\text{init}}$ in each layer yielded better performance.

\section{Experimental details}
\label{appendix:expdetails}

\subsection{EMPSN architecture and QM9}

For the data, we used the common split of $100\text{K}$ molecules for training, $10 \text{K}$ molecules for testing and the rest for validation. For both experiments, the models are trained for 1000 epochs each, where we used $~ 200\text{K}$ parameters for the small model comparison and $~ 1\text{M}$ parameters for the SOTA comparison. 
For the small model comparison, we used a $4$-layer EMPSN of dimension $2$ as our starting point. For the comparison where we compensate for the number of layers, we used $8$ layers in the $0$-dimensional model, and $6$ layers for the $1$ dimensional models. For the SOTA comparison, we used a $7$-layer EMPSN of dimension $2$.

The models are optimized using Adam \cite{kingma2014adam} with an initial learning rate of $\eta = 5 \cdot 10^{-4}$ and a Cosine Annealing learning rate scheduler \cite{loshchilov2016sgdr}. The loss used for optimization is the Mean Absolute Error. All predicted properties have been normalized by first subtracting the mean of the target in the training set and then dividing by the mean absolute deviation in the training set to stabilize training. We used a batch size of $128$ molecules per batch and a weight decay of $10^{-16}$. Last, we endowed the message and update functions with batch normalization. 

\subsection{N-body system}

For the data, we used the same setup as used in \citet{satorras2021n}, i.e. we used $3,000$ training trajectories, $2,000$ validation trajectories, and $2,000$ test trajectories, where each trajectory contains $1,000$ time steps. We used a $4$-layer EMPSN of dimension 2 for our experiments. 

The optimization is done using Adam, with a constant learning rate of $\eta = 5 \cdot 10^{-4}$, a batch size of $100$, and weight decay of $10^{-12}$. Moreover, the invariant features are embedded using Gaussian Fourier features as introduced in \citet{tancik2020fourier}. The loss minimized is the MSE in the predicted position.


\end{document}